\pdfoutput=1

\documentclass[11pt]{article}

\usepackage[final]{acl}

\usepackage{times}
\usepackage{latexsym}
\usepackage{booktabs}
\usepackage{multirow}
\usepackage{multicol}
\usepackage{amsmath}
\usepackage{float}
\usepackage{stfloats}

\usepackage[T1]{fontenc}

\usepackage[utf8]{inputenc}

\usepackage{microtype}

\usepackage{inconsolata}

\usepackage{graphicx}

\usepackage{bbold}
\usepackage{booktabs}
\usepackage{dsfont}
\usepackage{caption}
\usepackage{subcaption}
\usepackage{xspace}

\usepackage{amsmath,amsfonts,bm}

\def\eqref#1{equation~\ref{#1}}

\def\1{\bm{1}}

\DeclareMathAlphabet{\mathsfit}{\encodingdefault}{\sfdefault}{m}{sl}
\SetMathAlphabet{\mathsfit}{bold}{\encodingdefault}{\sfdefault}{bx}{n}

\definecolor{lightgray}{HTML}{808080}
\definecolor{perfup}{HTML}{008000}
\definecolor{perfdown}{HTML}{1d7b21}
\definecolor{myred}{HTML}{e60035}

\newcommand\redsout{\bgroup\markoverwith{\textcolor{red}{\rule[0.5ex]{2pt}{2pt}}}\ULon}
\newcommand\bluesin{\bgroup\markoverwith{\textcolor{green}{\rule[-0.5ex]{2pt}{2pt}}}\ULon}
\newcommand\redsin{\bgroup\markoverwith{\textcolor{red}{\rule[-0.5ex]{2pt}{2pt}}}\ULon}

\renewcommand{\sectionautorefname}{\S\kern-0.2em}
\renewcommand{\subsectionautorefname}{\S\kern-0.2em}

\definecolor{lightgreen}{HTML}{1d7b21}

\newcommand{\textpr}[1]{\textcolor{violet!70!black}{\texttt{\footnotesize #1}}}

\renewcommand{\vec}[1]{\bm{#1}}
\newcommand{\question}{\vec q}
\newcommand{\answer}{\vec a}
\newcommand{\explanation}{\vec e}

\newcommand{\model}{M}
\newcommand{\dpomodel}{M_{O}}
\newcommand{\studentmodel}{M_{S}}
\newcommand{\teachermodel}{M_{T}}
\newcommand{\response}{R}
\newcommand{\prompt}{Q}
\newcommand{\Lagr}{\mathcal{L}}
\newcommand{\ourname}{PEX\xspace}
\newcommand{\ournamelong}{\textbf{P}rediction-\textbf{EX}planation consistency\xspace}

\title{A Necessary Step toward Faithfulness:\\Measuring and Improving Consistency in Free-Text Explanations}

\author{Lingjun Zhao \\
  University of Maryland \\
  College Park, Maryland, USA \\
  \texttt{lzhao123@umd.edu} \\\And
  Hal Daum\'e III \\
  University of Maryland \\
  College Park, Maryland, USA \\
  \texttt{hal3@umd.edu} \\}

\begin{document}
\maketitle
\begin{abstract}
Faithful free-text explanations are important to ensure transparency in high-stakes AI decision-making contexts, but they are challenging to generate by language models and assess by humans.
In this paper, we present a measure for Prediction-EXplanation (\ourname) consistency, by extending the concept of weight of evidence. 
This measure quantifies how much a free-text explanation supports or opposes a prediction, serving as an important aspect of explanation faithfulness.
Our analysis reveals that more than 62\% explanations generated by large language models lack this consistency.
We show that applying direct preference optimization improves the consistency of generated explanations across three model families, with improvement ranging from 43.1\% to 292.3\%.
Furthermore, we demonstrate that optimizing this consistency measure can improve explanation faithfulness by up to 9.7\%.\footnote{\,Our code is
publicly released at \url{https://github.com/lingjunzhao/PEX_consistency}.}

\end{abstract}

\section{Introduction} \label{introduction}

Explainable AI systems are those that can specify the relationship between an output or prediction and the deductive or nomological process that led to that prediction through logically consistent and empirically grounded means \cite{woodward2003scientific}.\footnote{\,Other forms of explanations exist, such as those that aim to 
provide useful information for better decision making \cite{kayser-etal-2024-fool, han2023ignoranceblissrolepost} or debugging  \cite{ribeiro2016should, lundberg2017unified}.} Such explanations are often called ``faithful'' in the sense that they accurately reflect a model's true reasoning process \cite{jacovi-goldberg-2020-towards, lyu2024towards}, and faithfulness is often seen as crucial for trustworthiness, transparency, and accountability.

Faithfulness, however, is a difficult construct to measure for complex models (such as deep neural networks) precisely because we do not know a priori exactly how a models combines its input features to make a prediction, beyond trivially writing out the entire computation---which is then impenetrable to a person. When an explanation is given in the form of natural language, increasingly common especially in the context of large language models \cite{camburu2018snli, wiegreffe-etal-2022-reframing} the challenge increases because humans often conflate faithfulness and plausibility: how convincing it appears \cite{jacovi-goldberg-2020-towards, lyu-etal-2023-faithful, wiegreffe-etal-2022-reframing}. This leads to models that produce plausible yet unfaithful accounts of their reasoning processes \cite{ye2022unreliability, lanham2023measuringfaithfulnesschainofthoughtreasoning}, which risks eroding the very trustworthiness that they aim to improve.

\begin{figure}[t]
    \centering
    \includegraphics[width=.46\textwidth]{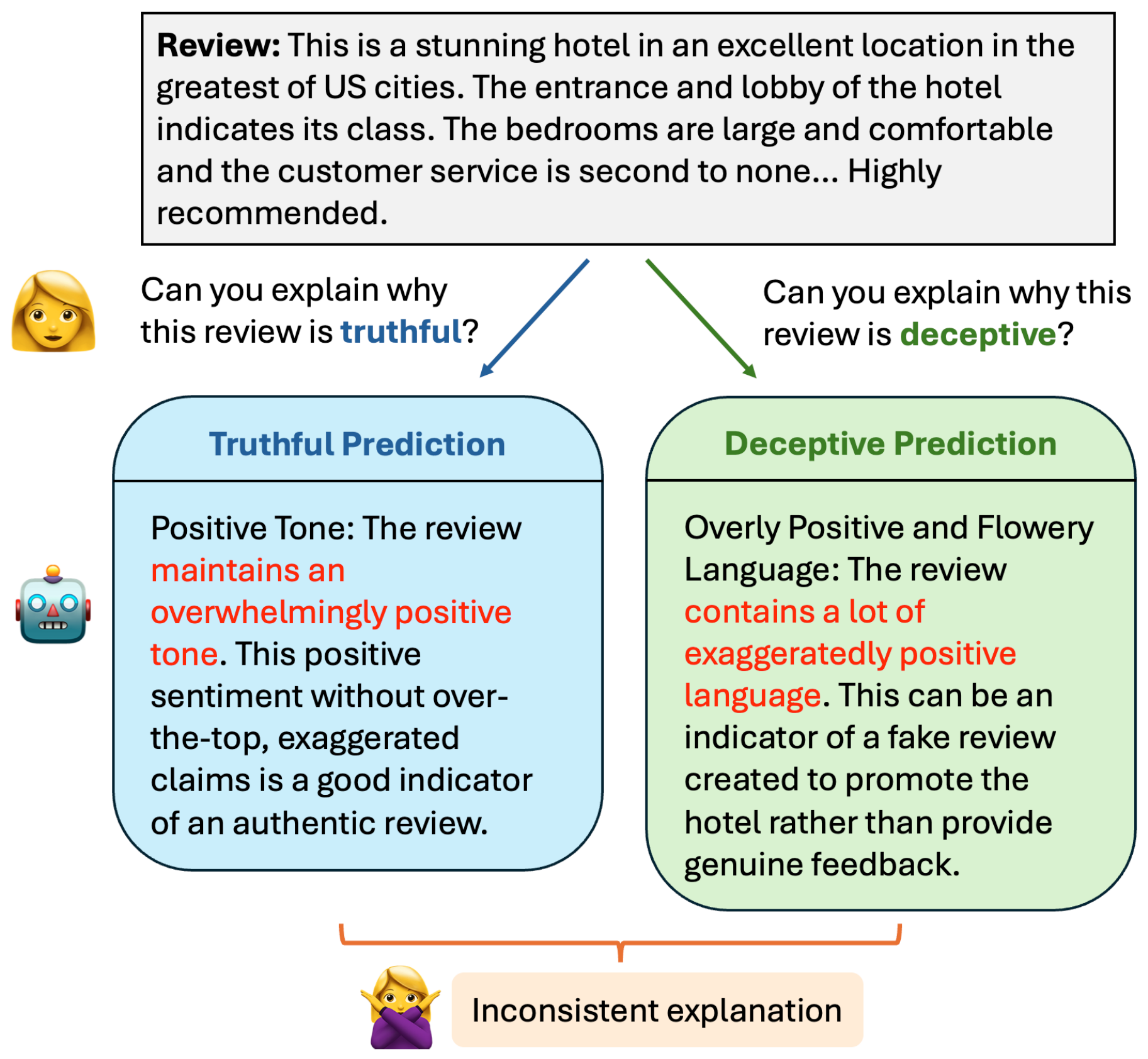} 
    \caption{Example of GPT-4 model generating explanations for  truthful or deceptive prediction about a hotel review’s authenticity. While each explanation appears plausible on its own, they are \textit{inconsistent}: both the truthful and deceptive explanations rely on the same evidence ``use a lot of positive language''.}
    \label{fig:teaser}
\end{figure}


In this paper, we aim to improve model \textit{faithfulness} by optimizing the \textit{consistency} between an explanation and its explanandum (e.g., a model prediction). Following \citet{miller2019explanation}'s criterion, a necessary condition for faithful explanations is that an explanation should contrastively refute the negation of the explanandum. Therefore, for an explanation to faithfully represent a deductive or nomological process, it cannot simultaneously support a prediction $y$ and its negation $\lnot y$. We formalize this notion as \textit{inconsistency}.
For example, in \autoref{fig:teaser}, a model uses essentially the same explanation (``use a lot of positive language') to argue in favor of a hotel review being both authentic and deceptive in this opinion spam classification task.

Our work proceeds in three steps. First, we ask how consistent large language model (LLM)-generated explanations are in practice. To do this, we introduce a measure of consistency, \ourname (for \ournamelong), which can be applied to models that provide probabilities over their generated outputs. 
\ourname leverages the weight of evidence framework \cite{melis2021human, wod1985weight} to quantify how much an explanation speaks in favor of (vs against) a prediction (\autoref{sec:problem}, \autoref{sec:consistency}). On two datasets---TripAdvisor hotel review and Amazon product review spam detection \cite{ott2013negative, hussain2020spam}---we find that 62\%–86\% of explanations from Llama-2 \cite{touvron2023llama}, Mistral \cite{mistral7b-instruct-v0.3}, and Yi-1.5 \cite{young2024yi} are inconsistent, logically implying that they are also unfaithful (\autoref{sec:exp_consistent}).

Second, we ask whether we can use our consistency measure \ourname to train a model to produce more consistent---and therefore less unfaithful---explanations. 
We show how to apply both supervised fine-tuning and direct preference optimization (DPO) \cite{rafailov2024direct} to refine pretrained LLMs (\autoref{sec:dpo}). 
For DPO, we sample explanations from LLMs and rank them according to \ourname; those that score highly are treated as ``preferred'' in the optimization, and those that score lowly are treated as ``dispreferred.'' 
Experimentally, we show this DPO-based approach significantly improves on supervised fine-tuning and can improve explanation consistency by 43.1\%-292.3\% (\autoref{sec:exp_dpo}). 

It is, of course, possible that the fine-tuning leads to more consistent explanations, but not more faithful explanations. And so finally we measure whether the explanations that we optimized for consistency lead to improved faithfulness. We construct a proxy measure of faithfulness by assuming that a more faithful explanation should be more useful for an external observer to predict a model's behavior \cite[see Section 3.4.2]{lyu2024towards}. Using this idea, we adopt a simulatability-based explanation faithfulness evaluation method \cite{pruthi2022evaluating} (\autoref{sec:faithfulness_eval}) and show that optimizing for \ourname improves faithfulness by 1.5\% to 9.7\% (\autoref{sec:exp_eval}).

\section{Related Work} \label{related_work}

\paragraph{Challenges in ensuring faithfulness of free-text explanations.}
Different from interpretable AI \cite{Wallace2018InterpretingNN, selvaraju2017grad, ribeiro2016should, briakou2023explaining}, 
generating explanations is a procedure to explicitly explaining model decisions to people \cite{miller2019explanation}.
A faithful explanation should accurately reflect the reasoning process behind the model's prediction \cite{jacovi-goldberg-2020-towards, lyu2024towards, ribeiro2016should}.
For models that first make a prediction with a standard black-box
predictor and then justify the prediction with an explainer \cite{camburu2018snli, park2018multimodal, wu-mooney-2019-faithful}, 
there is no guarantee for the explanation faithfulness. 
For models that first generate an explanation and then provided as the only input to the predictor, the explanations can still be self-inconsistent \cite{camburu-etal-2020-make, zhou-etal-2023-flame}, or optimized in terms of plausibility instead of faithfulness \cite{kumar-talukdar-2020-nile}.
Most methods that jointly explain and make prediction \cite{rajani-etal-2019-explain, narang2020wt5, ling-etal-2017-program, jung-etal-2022-maieutic, ramnath2024tailoring} also do not ensure faithfulness, as models may ignore explanation during prediction. 
For chain-of-thought style prompting methods \cite{wei2022chain, wang2022self, zhou2022least}, the explanations can be unfaithful \cite{turpin2024language, lanham2023measuringfaithfulnesschainofthoughtreasoning}. 
\citet{wiegreffe-etal-2022-reframing, Marasovi2021FewShotSW} show potential for generating plausible free-text explanations with only a few examples, but the explanations can be unfaithful \cite{ye2022unreliability, chen2024towards}. 

\paragraph{Evaluate explanation faithfulness.}

We do not assume access to ground-truth explanations for evaluation, reflecting real-world scenarios; therefore white-box faithfulness evaluation \cite{zhou2022feature, chen2018learning} is not applicable to our problem.
While robustness \cite{alvarez2018robustness} and perturbation-based evaluation methods \cite{samek2016evaluating} have been applied in the vision domain, their applicability to NLP is limited due to the discrete nature of language inputs, and removing a single word can render a sentence meaningless \cite{lyu2024towards}.
Simulatability-based methods have been employed to evaluate explanation \cite{hase-bansal-2020-evaluating, hase-etal-2020-leakage, doshi2017towards}. This method is recommended to evaluate faithfulness \cite{lyu2024towards}, as the more faithful an explanation is, the more information it should contain about the model’s decision mechanism, and thus the easier it would be for an external simulator, to predict the model’s behavior based on the explanation. The assumption is that if an explanation leads to a different prediction than that made by the model it explains, then it is unfaithful \cite{jacovi-goldberg-2020-towards}.
\citet{pruthi2022evaluating} propose a framework for evaluating explanations by measuring accuracy gains in a student model trained to mimic a teacher. 
This framework is designed for system-level evaluation rather than individual-level, as the student model requires more than one examples to learn from.


\section{Measuring and Improving Prediction-Explanation  Consistency}
\label{model}
\begin{figure*}[t!]
\centering
\includegraphics[width=\textwidth]{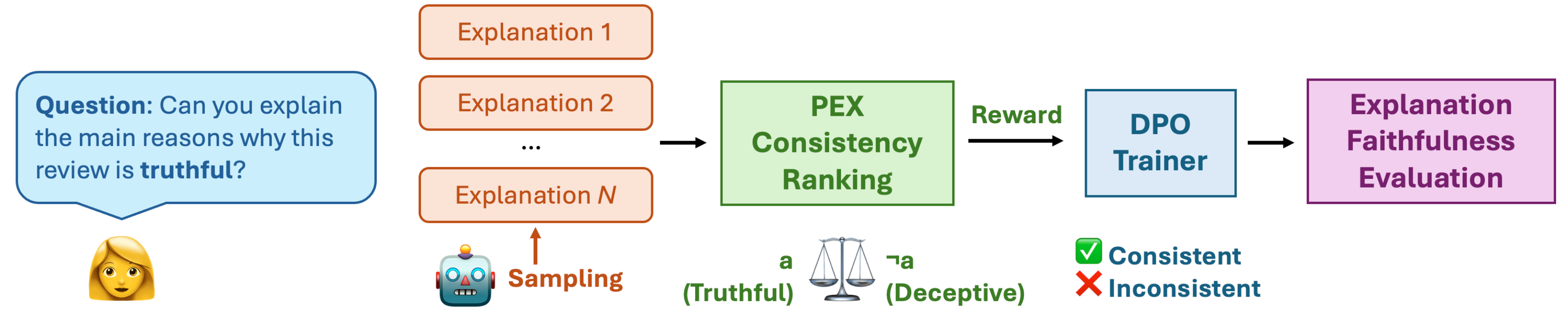}
\caption{Our framework for generating more consistent explanations. Given a question and a model predicted answer, we sample explanations from a language model and rank them using \ournamelong (\ourname), which measures how well an explanation $\explanation$ supports a given prediction $\answer$ compared to its negation $\lnot \answer$. This measure serves as a reward signal to construct preference dataset for direct preference optimization (DPO), improving \ourname consistency. Finally, we evaluate whether the consistency-optimized explanations are more faithful.}
\label{fig:pipeline}
\end{figure*}

Faithful explanations that accurately reflect a model’s reasoning \cite{jacovi-goldberg-2020-towards, lyu2024towards} 
promotes AI transparency and accountability. In this work, we aim to improve the faithfulness of free-text explanations by optimizing their consistency.
To this end, we first introduce \ourname (\ournamelong) measure to quantify how well an explanation speaks in favor of (vs against) a prediction (\autoref{sec:consistency}). We then use this measure as an optimization criterion to generate more consistent explanations (\autoref{sec:dpo}), as illustrated in \autoref{fig:pipeline}.
Finally, we evaluate the efficacy of this optimization in generating more faithful explanations (\autoref{sec:faithfulness_eval}).

\subsection{Problem: Inconsistent Explanations}
\label{sec:problem}

Our first goal is to develop a measure of whether a model’s explanations are \textit{consistent} in justifying its predictions $\answer$ over an its negation $\lnot \answer$, as a \textit{faithful} explanation cannot simultaneously support both predictions \cite{lipton1990contrastive, miller2019explanation, brassard2024acorn}.
We observe that language model generated explanations can be inconsistent,
e.g. in \autoref{fig:teaser}, 
the explanation ``use a lot of positive language'' supports both the truthful and deceptive hypotheses, failing to distinguish why the model predicted answer $\answer$ instead of the alternative prediction $\lnot \answer$.
As a result, the explanation is not consistent and, therefore, not faithful.
We focus on generating explanations to justify \textit{opinion spam detection} \cite{ott2011finding, hussain2020spam}.
We use this task because it requires minimal external knowledge while allowing for the generation of meaningful explanations.

We focus on language models that can answer questions, where a model $M$ takes a question $\question$ as input and generates an answer $\answer$ as output with probability $M(\answer \mid \question)$. 
Additionally, we assume that upon request (e.g. prompting or other mechanisms) that the same model $\model$ can generate a potentially faithful explanation of its prediction. We denote this as $\model(\explanation \mid \prompt(\question, \answer))$, where $\prompt$ is an appropriate prompt, and explanation $\explanation = (e_1, e_2, ..., e_n)$ is a sequence of words.
We use a one-shot example to prompt the model to generate explanation for its prediction. The explanation usually contains 2-3 rationales. The explanations in the one-shot example are generated by GPT-4 (detailed in \autoref{app:explanation_prompting}).

For example, in opinion spam detection, 
question $\question$ is formatted as: $\question=$ \textpr{Is this  review truthful or deceptive? Review: \{review\}}. The output $\answer$ is \textpr{Truthful} or \textpr{Deceptive}.
If $\model$ predicts the answer $\answer$ as \textpr{Truthful}, the prompt $\prompt$ is formatted as:
$\prompt(\question, \textpr{Truthful})=$ \textpr{Is this review truthful or deceptive? Review: \{review\}. Answer: Truthful. Question: Can you explain the main reasons why the review is truthful?}
If $\model$ predicts the answer $\answer$ as \textpr{Deceptive}, the prompt $\prompt$ is formatted as:
$\prompt(\question, \textpr{Deceptive})=$ \textpr{Is this review truthful or deceptive? Review: \{review\}. Answer: Deceptive. Question: Can you explain the main reasons why the review is deceptive?}
An example of the generated explanation $\explanation$ is \textpr{No specific examples: The review does not provide specific examples of the poor service or quality of the hotel.}

\subsection{Measuring Prediction-Explanation (\ourname) Consistency}
\label{sec:consistency}

We define explanation inconsistency as failing to explicate why the model predicted $\answer$ instead of its negation $\lnot \answer$.
To quantify this, we introduce \ournamelong\ (\ourname), which measures how well an explanation supports a given prediction compared to its negation. We build on the \textit{weight of evidence} framework \cite{melis2021human, wod1985weight}, extending it to to compute probabilities when evidence is given as sequences of words. The weight of evidence measures the extent to which an explanation supports or undermines a hypothesis, making it a reasonable measure of explanation consistency.

\paragraph{\ourname consistency.} To measure the \ourname consistency of explanation $\explanation$ for a given question $\question$ and model prediction $\answer$, we compute a score to compare the likelihood of model $\model$ generating $\explanation$ under different predictions, 
thereby quantifying how well the explanation supports the given prediction $\answer$ over its negation $\lnot \answer$:

\begin{equation}
   C(\explanation) = \log \frac{\model(\explanation \mid \prompt(\question, \answer))}{\model(\explanation \mid \prompt(\question, \lnot \answer))}
\end{equation}

\noindent  where the text prompt $\prompt$ and explanation generation $\model(\explanation \mid \prompt(\question, \answer))$ are formatted in \autoref{sec:problem}.
We estimate the conditional probability $\model(\explanation \mid \prompt(\question, \answer))$ using the chain rule:

\begin{multline}
\label{eq:woe}
    \model(\explanation \mid \prompt(\question, \answer))) = \model(e_1 \mid \prompt(\question, \answer))) \cdot \\
    \model(e_2 \mid \prompt(\question, \answer)), e_1) \cdot \\ \cdots \model(e_n \mid \prompt(\question, \answer)), e_1, \ldots, e_{n-1})
\end{multline}

\noindent  where $e_i$ is the $i$-th word of the explanation $\explanation$.
We estimate the conditional probability $\model(\explanation \mid \prompt(\question, \lnot \answer))$ using the same approach.

If we view $\model(\explanation \mid \prompt(\question, \answer))$ as $\model(\explanation \mid \question, \answer)$ by dropping the text format $Q$, \ourname consistency $C(\explanation)$ can also be computed using Bayes' rule as:

\begin{multline}
    \log \frac{\model(\explanation \mid \question, \answer)}
              {\model(\explanation \mid \question, \lnot \answer)} \\
    = \log \frac{\model(\answer \mid \question, \explanation)}
              {\model(\lnot \answer \mid \question, \explanation)}
     - \log \frac{\model(\answer \mid \question)}
              {\model(\lnot \answer \mid \question)}
\end{multline}

\noindent where $\log \frac{\model(\answer \mid \question, \explanation)}{\model(\lnot \answer \mid \question, \explanation)}$ is the posterior log-odds ratio between prediction $\answer$ and $\lnot \answer$ conditioned on question $\question$ and explanation $\explanation$. 
$\log \frac{\model(\answer \mid \question)}{\model(\lnot \answer \mid \question)}$ is the prior log-odds ratio, where $\model(\answer \mid \question)$ is defined in \autoref{sec:problem}.
To compute $\model(\answer \mid \question, \explanation)$ using a language model $\model$, 
we format the prompt $\prompt'$ as: $\prompt' (\question, \explanation)$ = \textpr{Is this review truthful or deceptive? Review: \{review\}. Analysis: {$\explanation$}.} 
Then, we compute adjusted \ourname consistency $C' (\explanation)$ as:

\begin{equation}
\label{eq:adjusted_woe}
C' (\explanation) 
= \log \frac{\model(\answer \mid \prompt'(\question, \explanation))}{\model(\lnot \answer \mid \prompt'(\question, \explanation))} - \log \frac{\model(\answer \mid \question)}{\model(\lnot \answer \mid \question)}
\end{equation}

We use \textit{adjusted \ourname consistency} (\autoref{eq:adjusted_woe}) in our experiments because computing sequence probability for the original \ourname consistency $C(\explanation)$ (\autoref{eq:woe}), requires density estimation, which is often less reliable than computing classification probabilities for $C' (\explanation)$. Additionally, sequence probability is typically affected by sequence length.

\paragraph{Supervised fine-tuning.}

We assume access to language models that provide output probabilities e.g. $\model(\answer \mid \prompt'(\question, \explanation))$,
 to compute \ourname consistency (\autoref{eq:adjusted_woe}).
Pretrained language models, including LLama-2 and Mistral, do not perform well on opinion spam detection, with prediction accuracy close to random guessing. Thus we fine-tune the models on the training dataset using supervised fine-tuning training (SFT) to have better estimation of $\model(\answer \mid \question)$. This is achieved by using the maximum-likelihood estimation (MLE) objective:

\begin{equation}
\sum_{(q,a) \in D} \log \model(\answer \mid \question)
\end{equation}

\noindent where the training dataset $D$ consisting of question answer pairs in the form of $(\question, \answer)$. 
We use the fine-tuned models to measure \ourname consistency.

\subsection{Multiple Sampling for Generating More Consistent Explanations}
\label{sec:sampling}
The \ourname measure (\autoref{sec:consistency}) can then be applied to rank language model generated explanations. 
For each question $\question$ and the answer $\answer$ predicted by model $\model$, we sample explanations $\explanation$ from $\model$ using random sampling from the probability distribution $\model(\explanation \mid \prompt(\question, \answer))$. 
We compute \ourname consistency score (\autoref{eq:adjusted_woe}) for each explanation, and rank the explanations according to the score.

\subsection{Optimizing Explanation Consistency with Direct Preference Optimization}
\label{sec:dpo}

As the language models can generate inconsistent explanations for their predictions, we train models $\dpomodel$ using the direct preference optimization (DPO) objective \cite{rafailov2024direct} to generate more consistent explanations.
\paragraph{DPO training.} Given a question $\question$ and answer $\answer$, $\dpomodel$ is trained to increase the likelihood $\dpomodel(\explanation_w \mid \prompt(\question, \answer)))$ of generating consistent explanation $\explanation_w$ as \textit{preferred completion}, and decrease the likelihood $\dpomodel(\explanation_l \mid \prompt(\question, \answer)))$ of generating inconsistent explanation $\explanation_l$ as \textit{dispreferred completion}.
We use the language model $\model$ (\autoref{sec:consistency}) as base reference model, and use the DPO objective $\Lagr (\dpomodel; \model)$:

\begin{multline}
    \mathbb{E}_{(\question, \answer, \explanation_w, \explanation_l) \sim D_O} \Bigg[ 
     \log \sigma \Big(\beta \log \frac{\dpomodel(\explanation_l \mid \prompt(\question, \answer))}{\model(\explanation_l\mid \prompt(\question, \answer))}  \\
    - \beta \log \frac{\dpomodel(\explanation_w \mid \prompt(\question, \answer))}{\model(\explanation_w \mid \prompt(\question, \answer) )}\Big)
    \Bigg] 
\end{multline}

\noindent where the weights of $\dpomodel$ are initalized from the base reference model $\model$. 
$\sigma$ is the logistic function, and $\beta$ is a parameter controlling the deviation from the base reference model $\model$, set to 0.1. 

\paragraph{Preference dataset.} To construct preference dataset $D_{O}$ for DPO training, we select $\explanation_w$ and $\explanation_l$ from the explanations generated by the reference language model $\model$. 
For each question $\question$ and answer $\answer$ predicted by model $\model$, we sample and rank the explanations using \ourname consistency score (\autoref{eq:adjusted_woe}), as described in \autoref{sec:sampling}.
We consider explanations in the top $p\%$ are consistent, and those in the bottom $p\%$ are inconsistent.
For each question and the answer, we use the combinations of the  consistent and inconsistent explanations as preferred and dispreferred completions ($\explanation_w$, $\explanation_l$) to train DPO.

\section{Explanation Faithfulness Evaluation}
\label{sec:faithfulness_eval}

\begin{figure}[t]
    \centering
    \includegraphics[width=.48\textwidth,clip,trim=10 110 10 110]{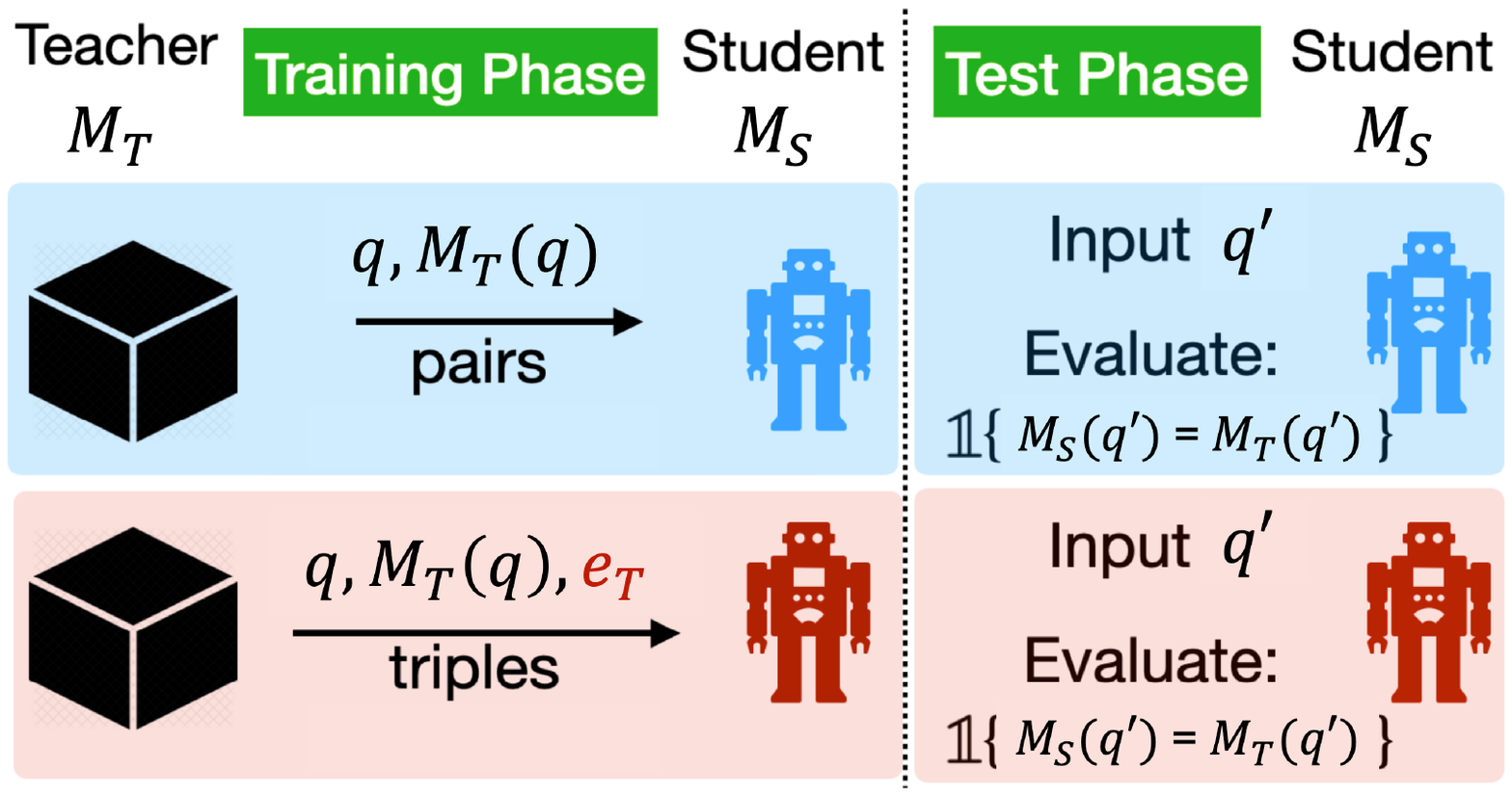} 
    \caption{Explanation evaluation framework \cite[figure reproduced from][]{pruthi2022evaluating}: Student models learn from a teacher without explanation (top) or with explanation (bottom) as side information. Explanations $\explanation_T$ are effective if they improve student performance to mimic teacher model's prediction on unseen examples during test phase without explanations (right).}
    \label{fig:evaluation}
\end{figure}

To evaluate the efficacy of optimizing \ourname consistency (\autoref{model}) in generating more faithful explanations,
we construct a proxy measure of faithfulness by assuming that a more faithful explanation should be more useful for an external observer to predict a model's behavior \cite[see Section 3.4.2]{lyu2024towards}.
Using this idea, we adopt a simulatability-based explanation faithfulness evaluation method \cite{pruthi2022evaluating},
as shown in \autoref{fig:evaluation}.
In this framework, a teacher model generates a prediction and explanation, and a student model is trained to simulate the teacher's model prediction. 
Explanations are available to the student model during training, but are not available during testing to avoid label leakage.
We use language models as student models for our evaluation. While human evaluation is valuable, it is prohibitively expensive at scale, and may be confounded by humans’ preconceived notions—especially when conflating explanation faithfulness with plausibility. While language models also reflect biases, they offer a consistent and scalable means of evaluation.

\paragraph{Student model training.} Our teacher model $\teachermodel$ is a supervised fine-tuned model $\model$ optimized for prediction accuracy (\autoref{sec:consistency}), and the student model $\studentmodel$ is initialized from a pretrained language model. 
We fine-tune the student model $\studentmodel$ to simulate the teacher's model prediction and explanation: $\studentmodel$ takes question $\question$ as input, and generates answer $\answer$ and explanation $\explanation$ autoregressively as its response, formatted as:
$\response(\answer, \explanation) = \textpr{Answer: $\answer$. Analysis: $\explanation$}$.
The model is trained to estimate $\studentmodel(\response(\answer, \explanation) \mid \question)$ using the MLE objective:

\begin{equation}
\sum_{(\question, \answer_T, \explanation_T) \in D_T} \: \sum_{n=1}^{N} log \studentmodel(\response_n \mid \response_{<n}, \question)
\end{equation}

\noindent where $\response_n$ is the $n$-th word of the text response $\response(\answer, \explanation)$, and $\response_{<n}$ is the first $n-1$ words of the response.
Dataset $D_T$ consisting of samples in the form of $(\question, \answer_T, \explanation_T)$, 
where $\answer_T = \teachermodel(\question)$ is the answer predicted by the teacher model $\teachermodel$, and explanation $\explanation_T$ is generated by $\teachermodel$ to justify its prediction $\answer_T$ to the question $\question$.
We generate the answer before the explanation for the student model, rather than after, because our preliminary experiments showed similar performance for both strategies. However, generating explanations at test time is computationally expensive. The answer-first strategy mitigates this overhead by enabling evaluation based on the likelihood of truthful or deceptive predictions without requiring explanation generation during inference.

\paragraph{Evaluation.} During testing, the student model generates prediction $\answer'=\studentmodel(\question')$ for  question $\question'$, and we measure the \textit{simulation performance} by comparing its prediction $\answer'$ with the teacher model's prediction $\answer_T'=\teachermodel(\question')$.

\section{Experimental Setup} \label{setup}

\paragraph{Datasets.}
We fine-tune pre-trained large language models to improve their prediction accuracy on two opinion spam detection datasets: (i) TripAdvisor hotel review dataset \cite{ott2013negative}, which contains 800 truthful reviews and 800 deceptive reviews. We randomly split the dataset to obtain 960 pairs of (review, label) for training, 320 pairs  for validation and 320 pairs for testing.  
(ii) Amazon product review dataset \cite{hussain2020spam}, which
we randomly select 1,000 truthful reviews and 1,000 deceptive reviews. We restrict reviews to those reviews containing at least 120 words to ensure that there is sufficient context for explanations. We split the selected reviews to obtain 1,200 pairs of (review, label) for training, 400 pairs for validation and 400 pairs for testing. 
We fine-tune the models on the train split, and select models on the validation split according to the F1 score. 

We generate explanations using the fine-tuned models on the validation split to compute \ourname consistency score statistics (\autoref{sec:consistency}). To evaluate explanation faithfulness, we also use the generated explanations on the validation split to serve as the teacher model to fine-tune a  student model (\autoref{sec:faithfulness_eval}). We report student model performance on the test split, evaluated against teacher predictions.

\paragraph{Pretrained-LLMs and SFT.}
We evaluate three different LLM architectures to generate explanations for the opinion spam detection datasets:
(i) Mistral-7B-Instruct-v0.3 \cite{mistral7b-instruct-v0.3},
(ii) Llama-2-13B-chat \cite{touvron2023llama},
and (iii) Yi-1.5-9B-chat \cite{young2024yi}.
We fine-tune the pretrained models (\autoref{sec:consistency}), achieving an F1 score of 94\% on the TripAdvisor validation set and 93\% on the Amazon validation set.
Details for fine-tuning the models are given in Appendix \autoref{app:sft}.

\paragraph{DPO training.}
To construct the preference dataset for DPO training (\autoref{sec:dpo}), we generate 40 explanations per review-prediction pair in the training split (\autoref{sec:sampling}) using a sampling temperature of 1.0. To increase diversity, we sample 40 additional explanations at a temperature of 1.2.
Explanations are ranked by their \ourname consistency score (\autoref{eq:adjusted_woe}), which measures how well they justify the prediction over its negation.
For TripAdvisor dataset, we randomly sample 8 explanation pairs: preferred completions are drawn from the top 10\% of scores and dispreferred from the bottom 10\%, excluding pairs where both scores are above or below zero to ensure contrast. For Amazon dataset, we sample 8 pairs from the top and bottom 5\% without applying a zero threshold, as many explanations for deceptive predictions have low scores.
After this process, we obtain 3,565 training samples on the TripAdvisor dataset for DPO training with the Llama-2 model, 2,796 samples for the Mistral model, and 4,083 samples for the Yi-1.5 model.
On the Amazon dataset, we obtain 9,600 training samples for each of the three language models.
Additional details on model training are provided in  \autoref{app:dpo}.

\paragraph{Student model training.}
We train student models\footnote{\,Our preliminary experiments do not suggest that using training-free in-context learning is fruitful.} using $k$ random examples from the validation split, where each example includes both the teacher model's prediction and its generated explanation (\autoref{sec:faithfulness_eval}).
We perform a five-pass training for $k=10$ and $k=20$.
We use small values of $k$ because the student model can learn effectively from the prediction label alone when $k$ is large.
The student model shares the same architecture as the teacher model but is \textit{not} fine-tuned on the training split.
Further details are provided in \autoref{app:sft}.

\paragraph{Evaluation metrics.}
To measure how well the explanations justifies the prediction over its negation, we compute \textbf{\ourname consistency} (\autoref{eq:adjusted_woe}) on model generated explanations using the corresponding supervised fine-tuned model. 
To assess explanation faithfulness, we evaluate the trained student models using \textbf{simulation performance (F1 score)} on the test split, ensuring teacher model's explanations are not provided as input to prevent label leakage (\autoref{sec:faithfulness_eval}). The simulation F1 score is computed by using the teacher model’s predictions as ground-truth labels.
We report average F1 score across all student model training passes.

\section{Experiments} \label{experiments}
We investigate the following questions:

\begin{enumerate}
\item How consistent are the explanations generated by large language models (LLM)?
\item Can the consistency of LLM-generated explanations be improved?
\item Are explanations optimized for consistency also more faithful?
\end{enumerate}

To address Q1, we analyze the \ourname consistency score distribution of the explanations. 
For Q2, we train models using direct preference optimization to generate more consistent explanations.
For Q3, we perform explanation faithfulness evaluation.

\subsection{How consistent are the explanations generated by large language models?}
\label{sec:exp_consistent}

\begin{figure}[h]
    \centering
    \includegraphics[width=.48\textwidth,clip,trim=5 5 5 10]
    {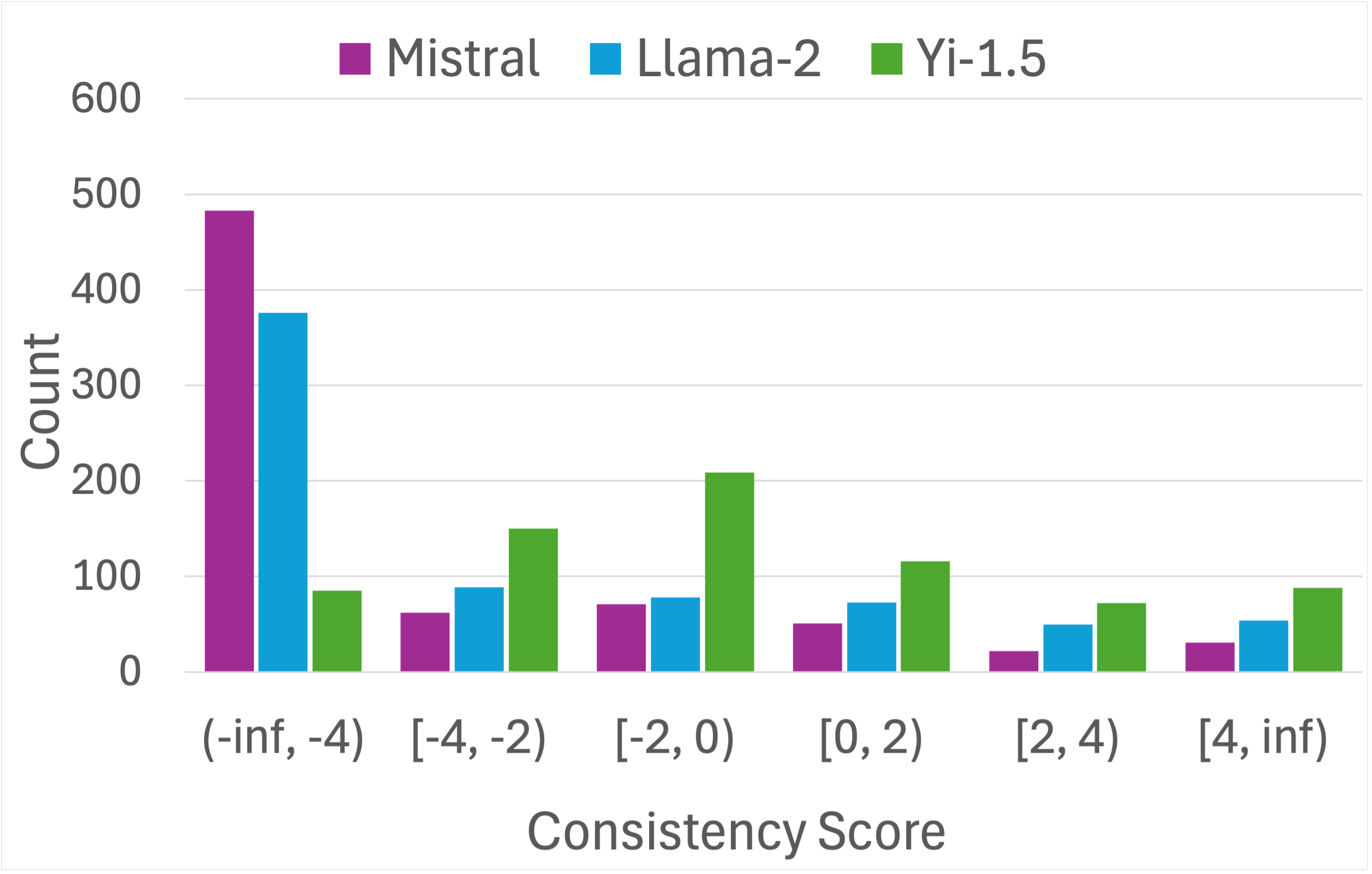} 
    \caption{\ournamelong (PEX) score distribution for different large language models. Explanations with PEX score higher than 2 are considered consistent \cite{wod1985weight}. \textbf{Takeaway:} models can generate 62\%-86\% inconsistent explanations.}
    \label{fig:woe_distribution}
\end{figure}

\paragraph{Consistency distribution.} \autoref{fig:woe_distribution} shows the distribution of \ourname consistency scores (\autoref{sec:consistency}) for explanations generated by the Mistral, Llama-2 and Yi-1.5 models on two opinion spam datasets.
According to \citet{wod1985weight, melis2021human}, weight of evidence scores above 2 are considered consistent. 
For the Mistral model, 
we see that 85.6\% of explanations are inconsistent (with \ourname score $<$ 0), 75.4\% for the Llama-2 model, and 61.7\% for the Yi-1.5 model.
Since inconsistent explanations do not support the model's predictions effectively, they pose a challenge for humans trying to understand how the model arrives at its decisions based on those explanations.

\paragraph{Sensitivity to prompt format.}  
To assess whether the PEX consistency score is sensitive to different prompt formats, we use 3 different prompt formats (detailed in \autoref{app:prompt_format}) to compute PEX consistency score and compute Kendall rank correlations between PEX consistency scores obtained from each pair of prompt versions. We find that PEX scores are relatively robust to variations in prompt format, as shown in \autoref{tab:sensitivity}. 

\begin{table}[]
\small
\centering
\begin{tabular}{lccc}
\toprule
\textbf{Model} & \textbf{V1 vs V2} & \textbf{V1 vs V3} & \textbf{V2 vs V3} \\
\midrule
\textbf{Llama-2} & (0.42--0.52)$^{\dagger}$ & (0.33--0.44)$^{\dagger}$ & (0.52--0.60)$^{\ddagger}$ \\
\textbf{Mistral} & (0.21--0.33)$^{\dagger}$ & (0.48--0.56)$^{\ddagger}$ & (0.38--0.48)$^{\dagger}$ \\
\textbf{Yi-1.5}  & (0.37--0.47)$^{\dagger}$ & (0.50--0.58)$^{\ddagger}$ & (0.40--0.50)$^{\dagger}$ \\
\bottomrule
\end{tabular}
\caption{Kendall rank correlations of PEX consistency scores across prompt versions on the validation set. Ranges indicate 90\% bootstrap confidence intervals. Correlations are moderate ($^{\dagger}$: mean $\geq$0.26) to strong ($^{\ddagger}$: mean $\geq$0.49). \textbf{Takeaway:} PEX scores are relatively robust to variations in prompt format.}
\label{tab:sensitivity}
\end{table}

\begin{table*}[!htbp]
    \centering
    \small
    \renewcommand{\arraystretch}{1.2} 
    \setlength{\tabcolsep}{8pt} 
\begin{tabular}{lccc|ccc}
    \toprule
    \multirow{2}{*}{\textbf{Model}} & \multicolumn{3}{c}{\textbf{TripAdvisor}} & \multicolumn{3}{c}{\textbf{Amazon}} \\
    \cmidrule(lr){2-4} \cmidrule(lr){5-7}
    & \textbf{Mistral} & \textbf{Llama-2} & \textbf{Yi-1.5} & \textbf{Mistral} & \textbf{Llama-2} & \textbf{Yi-1.5} \\
    \midrule
    \textbf{Pred Only} & 55.5 & 59.6 & 63.0 & 83.2 & 88.6 & 67.8 \\ \midrule
    \textbf{+ SFT}  & 66.2 & 58.3 & 66.0 & 85.2 & 89.2 & 75.5 \\
    \textbf{+ DPO}  & \textbf{69.8}$^{\dagger}$ & \textbf{63.4}$^{\dagger}$ & \textbf{70.9}$^{\dagger}$ & \textbf{86.7}$^{\dagger}$ & \textbf{91.6}$^{\dagger}$ & \textbf{85.2}$^{\dagger}$ \\
    \bottomrule
\end{tabular}
    \caption{Simulation performance (F1) of student models on the TripAdvisor and Amazon test set, evaluating how well they approximate the teacher model’s predictions on unseen examples. The student models are trained using explanations from different teacher models: supervised fine-tuning (SFT) or direct preference optimization (DPO) with \ourname consistency measure. $^{\dagger}$ indicates results that are significantly higher than those of the SFT model, with $p < 0.05$ as determined by a two-related-sample t-test. \autoref{app:eval} shows confidence intervals across training passes. \textbf{Takeaway:} optimizing \ourname consistency improves explanation faithfulness.}
    \label{tab:simulation_accuracy_dpo}
\end{table*}

\subsection{Can the consistency of LLM-generated explanations be improved?}
\label{sec:exp_dpo}

\begin{figure}[H]
\centering
\begin{subfigure}{0.42\textwidth}
\includegraphics[width=0.95\textwidth]
{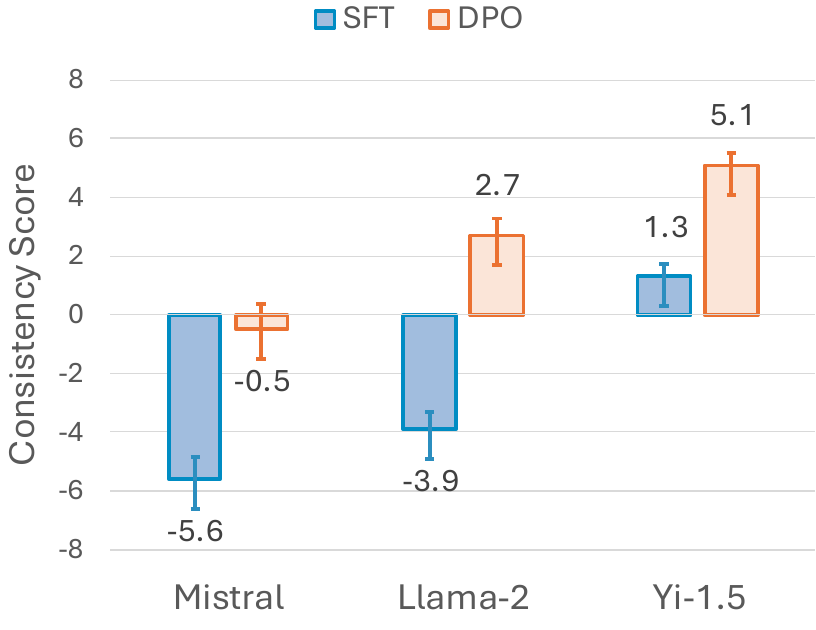}
\centering
         \caption{
         TripAdvisor dataset}
\end{subfigure}
\hfill \\
\begin{subfigure}{0.42\textwidth}
\includegraphics[width=0.95\textwidth]{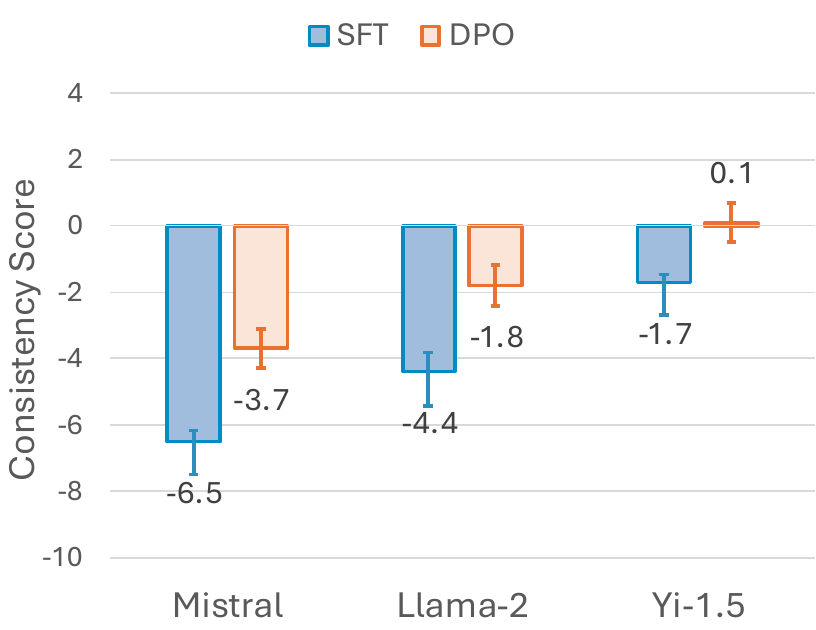}
\centering
         \caption{Amazon dataset}
\end{subfigure}
    \caption{Explanation average \ourname consistency scores for supervised fine-tuning (SFT) and direct preference optimization (DPO) models, computed on the TripAdvisor and Amazon validation datasets. The error bars represent 95\% confidence intervals. DPO model scores are significantly higher than those of the SFT model ($p < 0.05$ from two-related-sample t-test). \textbf{Takeaway:} DPO training can improve explanation consistency.}
    \label{fig:dpo}
\end{figure}

We use direct preference optimization (DPO) to enhance explanation consistency (\autoref{sec:dpo}), the results are illustrated in \autoref{fig:dpo}. 
For the Mistral model, DPO improves consistency by 2.8 to 5.1 points compared to explanations generated from the supervised fine-tuning (SFT) model (\autoref{sec:consistency}). DPO also improves 2.6 to 6.6 points for the Llama-2 model, and 1.8 to 3.8 points for the Yi-1.5 model. These results demonstrate the effectiveness of DPO in improving the \ourname consistency of the generated explanations, by learning the patterns that contribute to more consistent explanations.

For the Mistral model on the TripAdvisor dataset, the average consistency scores of DPO-generated explanations remain below zero. This arises from the upper bound imposed by the sampled explanations being used to construct preference data: the highest consistency score among DPO-sampled explanations for this model averages -2.1 across the reviews, which is lower than the Llama-2 model’s highest score of 4.4. 
Similarly, the highest consistency score among DPO-sampled explanations for Mistral model averages -2.8 on the Amazon dataset, which is lower than the Yi-1.5 model’s highest score of 0.3.

\subsection{Are explanations optimized for consistency also more faithful?}
\label{sec:exp_eval}
\begin{figure*}[h]
    \centering
    \includegraphics[width=\textwidth]{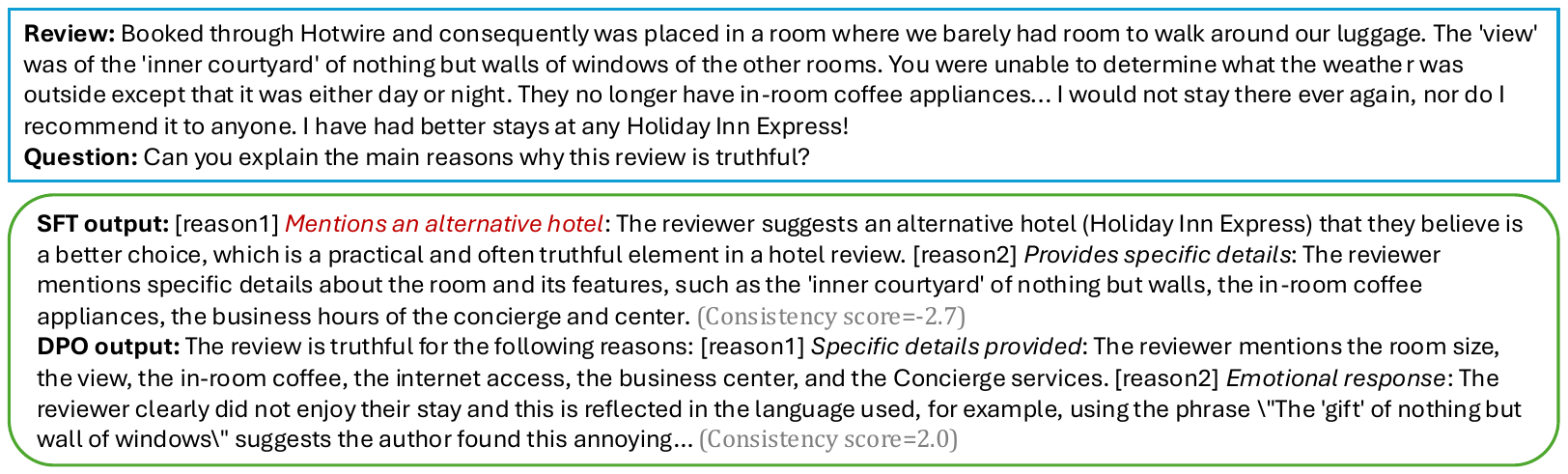} 
    \caption{
    Direct preference optimization (DPO) generates better explanations: The Mistral SFT (supervised fine-tuning) model predicts the TripAdvisor review as truthful; however, its explanation includes the phrase \textpr{mentions an alternative hotel}, which is often associated with deceptive prediction (e.g. \textpr{Comparison to other hotels: The reviewer mentions other hotels that are "better" or have "better service"}), leading to inconsistency. In contrast, the DPO model produces more faithful explanations with a higher consistency score.}
    \label{fig:qual_example1}
\end{figure*}

\paragraph{Systems.} We evaluate the faithfulness of explanations (\autoref{sec:faithfulness_eval}) generated by two systems:
(i) \textit{SFT}, a supervised fine-tuning model (\autoref{sec:consistency}), and
(ii) \textit{DPO}, a direct preference optimization model that improves \ourname consistency (\autoref{sec:dpo}).
Additionally, we report simulation performance for \textit{Pred Only}, which provides only the teacher model’s prediction without any explanations during student model training.
We use the same review across all three systems to enable a fair comparison.

\paragraph{Consistency-optimized explanations improve explanation faithfulness.}
\autoref{tab:simulation_accuracy_dpo} presents the student model’s simulation performance on the test set, where the student model was trained using explanations generated by different teacher models on the validation set.
The F1 score measures how well the student model approximates the teacher model’s predictions on unseen data, averaged over different number of training samples (\autoref{setup}).
On the TripAdvisor dataset, using DPO-generated explanations improves the F1 score by up to 14.3 points compared to not using explanations, and up to 5.1 points compared to using SFT explanations.
The results indicate that explanations optimized using the \ourname consistency measure with DPO better enable the student model to simulate the teacher’s predictions compared to explanations generated by the corresponding SFT model.
On the Amazon dataset, using DPO-generated explanations improves the F1 score by up to 17.4 points compared to not using explanations, and up to 9.7 points compared to using SFT explanations. The DPO explanations from the Yi-1.5 model result in a larger F1 score gap compared to the SFT explanations, likely due to the model’s higher average \ourname consistency score relative to the Mistral and Llama-2 models.

\paragraph{Qualitative examples.}

\autoref{fig:qual_example1} presents an example where the explanation generated by the DPO model receives a higher \ourname consistency score than the explanation generated by the SFT model. The SFT-generated explanation is also less faithful, as it often supports the deceptive prediction. Additional examples are provided in \autoref{app:qualitative}.

\section{Conclusion} \label{conclusion}

We present a new consistency measure for free-text explanations as an important aspect of faithfulness. We optimize explanation consistency using direct preference optimization, and show that it can improve the consistency and faithfulness of model generated explanations. 
We hope this research advances explanation faithfulness and enhances transparency in AI-assisted decision-making.
Another direction for future research is extending our approach to chain-of-thought explanations. Since our consistency measure and evaluation framework treat explanations and answers as independent variables, they are applicable regardless of how explanations are generated.

\section*{Limitations}
Our approach to improving explanation consistency relies on computing the sequence output’s conditional probability from language models. Consequently, this approach cannot be directly applied to explanations generated by completely black-box models.
We evaluate explanation faithfulness at the system level rather than at the individual explanation level, as the faithfulness evaluation framework we use is designed for system-level assessment.
Our experiments focus on enhancing explanations for binary classification, but we envision future work extending this approach to multi-class classification, as weight of evidence can be applied in that scenario as well \cite{melis2021human}.

\section*{Acknowledgements}
This material is based upon work supported by the NSF under Grant No. 2229885 (NSF Institute for Trustworthy AI in Law and Society, TRAILS), and U.S. Army Grant No. W911NF2120076.
We thank Marine Carpuat, Navita Goyal, Amanda Liu, Connor Baumler, Ilaria Canavotto, John Horty, Eric Pacuit for discussion on explanation roles and evaluation. We also thank Nguyen X. Khanh for suggestion on direct preference optimization training, and Mingyang Xie for writing feedback.

\bibliography{custom}

\begin{thebibliography}{52}
\providecommand{\natexlab}[1]{#1}

\bibitem[{Alvarez-Melis and Jaakkola(2018)}]{alvarez2018robustness}
David Alvarez-Melis and Tommi~S Jaakkola. 2018.
\newblock On the robustness of interpretability methods.
\newblock \emph{arXiv preprint arXiv:1806.08049}.

\bibitem[{Brassard et~al.(2024)Brassard, Heinzerling, Kudo, Sakaguchi, and Inui}]{brassard2024acorn}
Ana Brassard, Benjamin Heinzerling, Keito Kudo, Keisuke Sakaguchi, and Kentaro Inui. 2024.
\newblock \href {https://openreview.net/forum?id=2oHnsM9M9D} {{ACORN}: Aspect-wise commonsense reasoning explanation evaluation}.
\newblock In \emph{First Conference on Language Modeling}.

\bibitem[{Briakou et~al.(2023)Briakou, Goyal, and Carpuat}]{briakou2023explaining}
Eleftheria Briakou, Navita Goyal, and Marine Carpuat. 2023.
\newblock Explaining with contrastive phrasal highlighting: A case study in assisting humans to detect translation differences.
\newblock In \emph{Proceedings of the 2023 Conference on Empirical Methods in Natural Language Processing}, pages 11220--11237.

\bibitem[{Camburu et~al.(2018)Camburu, Rockt{\"a}schel, Lukasiewicz, and Blunsom}]{camburu2018snli}
Oana-Maria Camburu, Tim Rockt{\"a}schel, Thomas Lukasiewicz, and Phil Blunsom. 2018.
\newblock e-snli: Natural language inference with natural language explanations.
\newblock \emph{Advances in Neural Information Processing Systems}, 31.

\bibitem[{Camburu et~al.(2020)Camburu, Shillingford, Minervini, Lukasiewicz, and Blunsom}]{camburu-etal-2020-make}
Oana-Maria Camburu, Brendan Shillingford, Pasquale Minervini, Thomas Lukasiewicz, and Phil Blunsom. 2020.
\newblock \href {https://doi.org/10.18653/v1/2020.acl-main.382} {Make up your mind! adversarial generation of inconsistent natural language explanations}.
\newblock In \emph{Proceedings of the 58th Annual Meeting of the Association for Computational Linguistics}, pages 4157--4165, Online. Association for Computational Linguistics.

\bibitem[{Chen et~al.(2018)Chen, Song, Wainwright, and Jordan}]{chen2018learning}
Jianbo Chen, Le~Song, Martin Wainwright, and Michael Jordan. 2018.
\newblock Learning to explain: An information-theoretic perspective on model interpretation.
\newblock In \emph{International conference on machine learning}, pages 883--892. PMLR.

\bibitem[{Chen et~al.(2024)Chen, Singh, Liu, Zuo, Yu, He, and Gao}]{chen2024towards}
Yanda Chen, Chandan Singh, Xiaodong Liu, Simiao Zuo, Bin Yu, He~He, and Jianfeng Gao. 2024.
\newblock Towards consistent natural-language explanations via explanation-consistency finetuning.
\newblock \emph{arXiv preprint arXiv:2401.13986}.

\bibitem[{Doshi-Velez and Kim(2017)}]{doshi2017towards}
Finale Doshi-Velez and Been Kim. 2017.
\newblock Towards a rigorous science of interpretable machine learning.
\newblock \emph{arXiv preprint arXiv:1702.08608}.

\bibitem[{Good(1985)}]{wod1985weight}
I.~J. Good. 1985.
\newblock Weight of evidence: A brief survey.
\newblock \emph{Bayesian statistics}, 2:249--270.

\bibitem[{Han et~al.(2023)Han, Ektefaie, Farhat, Zitnik, and Lakkaraju}]{han2023ignoranceblissrolepost}
Tessa Han, Yasha Ektefaie, Maha Farhat, Marinka Zitnik, and Himabindu Lakkaraju. 2023.
\newblock \href {https://arxiv.org/abs/2312.05690} {Is ignorance bliss? the role of post hoc explanation faithfulness and alignment in model trust in laypeople and domain experts}.
\newblock \emph{Preprint}, arXiv:2312.05690.

\bibitem[{Hase and Bansal(2020)}]{hase-bansal-2020-evaluating}
Peter Hase and Mohit Bansal. 2020.
\newblock \href {https://doi.org/10.18653/v1/2020.acl-main.491} {Evaluating explainable {AI}: Which algorithmic explanations help users predict model behavior?}
\newblock In \emph{Proceedings of the 58th Annual Meeting of the Association for Computational Linguistics}, pages 5540--5552, Online. Association for Computational Linguistics.

\bibitem[{Hase et~al.(2020)Hase, Zhang, Xie, and Bansal}]{hase-etal-2020-leakage}
Peter Hase, Shiyue Zhang, Harry Xie, and Mohit Bansal. 2020.
\newblock \href {https://doi.org/10.18653/v1/2020.findings-emnlp.390} {Leakage-adjusted simulatability: Can models generate non-trivial explanations of their behavior in natural language?}
\newblock In \emph{Findings of the Association for Computational Linguistics: EMNLP 2020}, pages 4351--4367, Online. Association for Computational Linguistics.

\bibitem[{Hu et~al.(2021)Hu, Shen, Wallis, Allen-Zhu, Li, Wang, Wang, and Chen}]{hu2021lora}
Edward~J Hu, Yelong Shen, Phillip Wallis, Zeyuan Allen-Zhu, Yuanzhi Li, Shean Wang, Lu~Wang, and Weizhu Chen. 2021.
\newblock Lora: Low-rank adaptation of large language models.
\newblock \emph{arXiv preprint arXiv:2106.09685}.

\bibitem[{Hussain et~al.(2020)Hussain, Mirza, Hussain, Iqbal, and Memon}]{hussain2020spam}
Naveed Hussain, Hamid~Turab Mirza, Ibrar Hussain, Faiza Iqbal, and Imran Memon. 2020.
\newblock Spam review detection using the linguistic and spammer behavioral methods.
\newblock \emph{IEEE Access}, 8:53801--53816.

\bibitem[{Jacovi and Goldberg(2020)}]{jacovi-goldberg-2020-towards}
Alon Jacovi and Yoav Goldberg. 2020.
\newblock \href {https://doi.org/10.18653/v1/2020.acl-main.386} {Towards faithfully interpretable {NLP} systems: How should we define and evaluate faithfulness?}
\newblock In \emph{Proceedings of the 58th Annual Meeting of the Association for Computational Linguistics}, pages 4198--4205, Online. Association for Computational Linguistics.

\bibitem[{Jung et~al.(2022)Jung, Qin, Welleck, Brahman, Bhagavatula, Le~Bras, and Choi}]{jung-etal-2022-maieutic}
Jaehun Jung, Lianhui Qin, Sean Welleck, Faeze Brahman, Chandra Bhagavatula, Ronan Le~Bras, and Yejin Choi. 2022.
\newblock \href {https://doi.org/10.18653/v1/2022.emnlp-main.82} {Maieutic prompting: Logically consistent reasoning with recursive explanations}.
\newblock In \emph{Proceedings of the 2022 Conference on Empirical Methods in Natural Language Processing}, pages 1266--1279, Abu Dhabi, United Arab Emirates. Association for Computational Linguistics.

\bibitem[{Kayser et~al.(2024)Kayser, Menzat, Emde, Bercean, Novak, Morgado, Papiez, Gaube, Lukasiewicz, and Camburu}]{kayser-etal-2024-fool}
Maxime~Guillaume Kayser, Bayar Menzat, Cornelius Emde, Bogdan~Alexandru Bercean, Alex Novak, Abdal{\'a} Trinidad~Espinosa Morgado, Bartlomiej Papiez, Susanne Gaube, Thomas Lukasiewicz, and Oana-Maria Camburu. 2024.
\newblock \href {https://doi.org/10.18653/v1/2024.emnlp-main.1051} {Fool me once? {Contrasting} textual and visual explanations in a clinical decision-support setting}.
\newblock In \emph{Proceedings of the 2024 Conference on Empirical Methods in Natural Language Processing}, pages 18891--18919, Miami, Florida, USA. Association for Computational Linguistics.

\bibitem[{Kumar and Talukdar(2020)}]{kumar-talukdar-2020-nile}
Sawan Kumar and Partha Talukdar. 2020.
\newblock \href {https://doi.org/10.18653/v1/2020.acl-main.771} {{NILE} : Natural language inference with faithful natural language explanations}.
\newblock In \emph{Proceedings of the 58th Annual Meeting of the Association for Computational Linguistics}, pages 8730--8742, Online. Association for Computational Linguistics.

\bibitem[{Lanham et~al.(2023)Lanham, Chen, Radhakrishnan, Steiner, Denison, Hernandez, Li, Durmus, Hubinger, Kernion, Lukošiūtė, Nguyen, Cheng, Joseph, Schiefer, Rausch, Larson, McCandlish, Kundu, Kadavath, Yang, Henighan, Maxwell, Telleen-Lawton, Hume, Hatfield-Dodds, Kaplan, Brauner, Bowman, and Perez}]{lanham2023measuringfaithfulnesschainofthoughtreasoning}
Tamera Lanham, Anna Chen, Ansh Radhakrishnan, Benoit Steiner, Carson Denison, Danny Hernandez, Dustin Li, Esin Durmus, Evan Hubinger, Jackson Kernion, Kamilė Lukošiūtė, Karina Nguyen, Newton Cheng, Nicholas Joseph, Nicholas Schiefer, Oliver Rausch, Robin Larson, Sam McCandlish, Sandipan Kundu, Saurav Kadavath, Shannon Yang, Thomas Henighan, Timothy Maxwell, Timothy Telleen-Lawton, Tristan Hume, Zac Hatfield-Dodds, Jared Kaplan, Jan Brauner, Samuel~R. Bowman, and Ethan Perez. 2023.
\newblock \href {https://arxiv.org/abs/2307.13702} {Measuring faithfulness in chain-of-thought reasoning}.
\newblock \emph{Preprint}, arXiv:2307.13702.

\bibitem[{Ling et~al.(2017)Ling, Yogatama, Dyer, and Blunsom}]{ling-etal-2017-program}
Wang Ling, Dani Yogatama, Chris Dyer, and Phil Blunsom. 2017.
\newblock \href {https://doi.org/10.18653/v1/P17-1015} {Program induction by rationale generation: Learning to solve and explain algebraic word problems}.
\newblock In \emph{Proceedings of the 55th Annual Meeting of the Association for Computational Linguistics (Volume 1: Long Papers)}, pages 158--167, Vancouver, Canada. Association for Computational Linguistics.

\bibitem[{Lipton(1990)}]{lipton1990contrastive}
Peter Lipton. 1990.
\newblock Contrastive explanation.
\newblock \emph{Royal Institute of Philosophy Supplements}, 27:247--266.

\bibitem[{Lundberg and Lee(2017)}]{lundberg2017unified}
Scott~M Lundberg and Su-In Lee. 2017.
\newblock A unified approach to interpreting model predictions, nov.
\newblock \emph{NeurIPS}.

\bibitem[{Lyu et~al.(2024)Lyu, Apidianaki, and Callison-Burch}]{lyu2024towards}
Qing Lyu, Marianna Apidianaki, and Chris Callison-Burch. 2024.
\newblock Towards faithful model explanation in {NLP}: A survey.
\newblock \emph{Computational Linguistics}, pages 1--67.

\bibitem[{Lyu et~al.(2023)Lyu, Havaldar, Stein, Zhang, Rao, Wong, Apidianaki, and Callison-Burch}]{lyu-etal-2023-faithful}
Qing Lyu, Shreya Havaldar, Adam Stein, Li~Zhang, Delip Rao, Eric Wong, Marianna Apidianaki, and Chris Callison-Burch. 2023.
\newblock \href {https://doi.org/10.18653/v1/2023.ijcnlp-main.20} {Faithful chain-of-thought reasoning}.
\newblock In \emph{Proceedings of the 13th International Joint Conference on Natural Language Processing and the 3rd Conference of the Asia-Pacific Chapter of the Association for Computational Linguistics (Volume 1: Long Papers)}, pages 305--329, Nusa Dua, Bali. Association for Computational Linguistics.

\bibitem[{Marasovi{\'c} et~al.(2021)Marasovi{\'c}, Beltagy, Downey, and Peters}]{Marasovi2021FewShotSW}
Ana Marasovi{\'c}, Iz~Beltagy, Doug Downey, and Matthew~E. Peters. 2021.
\newblock \href {https://api.semanticscholar.org/CorpusID:244130199} {Few-shot self-rationalization with natural language prompts}.
\newblock In \emph{NAACL-HLT}.

\bibitem[{Melis et~al.(2021)Melis, Kaur, Daum{\'e}~III, Wallach, and Vaughan}]{melis2021human}
David~Alvarez Melis, Harmanpreet Kaur, Hal Daum{\'e}~III, Hanna Wallach, and Jennifer~Wortman Vaughan. 2021.
\newblock From human explanation to model interpretability: A framework based on weight of evidence.
\newblock In \emph{Proceedings of the AAAI Conference on Human Computation and Crowdsourcing}, volume~9, pages 35--47.

\bibitem[{Miller(2019)}]{miller2019explanation}
Tim Miller. 2019.
\newblock Explanation in artificial intelligence: Insights from the social sciences.
\newblock \emph{Artificial intelligence}, 267:1--38.

\bibitem[{Mistral(2023)}]{mistral7b-instruct-v0.3}
Mistral. 2023.
\newblock Mistral-7b-instruct-v0.3.
\newblock \url{https://huggingface.co/mistralai/Mistral-7B-Instruct-v0.3}.

\bibitem[{Narang et~al.(2020)Narang, Raffel, Lee, Roberts, Fiedel, and Malkan}]{narang2020wt5}
Sharan Narang, Colin Raffel, Katherine Lee, Adam Roberts, Noah Fiedel, and Karishma Malkan. 2020.
\newblock Wt5?! training text-to-text models to explain their predictions.
\newblock \emph{arXiv preprint arXiv:2004.14546}.

\bibitem[{Ott et~al.(2013)Ott, Cardie, and Hancock}]{ott2013negative}
Myle Ott, Claire Cardie, and Jeffrey~T Hancock. 2013.
\newblock Negative deceptive opinion spam.
\newblock In \emph{Proceedings of the 2013 conference of the north american chapter of the association for computational linguistics: human language technologies}, pages 497--501.

\bibitem[{Ott et~al.(2011)Ott, Choi, Cardie, and Hancock}]{ott2011finding}
Myle Ott, Yejin Choi, Claire Cardie, and Jeffrey~T Hancock. 2011.
\newblock Finding deceptive opinion spam by any stretch of the imagination.
\newblock \emph{arXiv preprint arXiv:1107.4557}.

\bibitem[{Park et~al.(2018)Park, Hendricks, Akata, Rohrbach, Schiele, Darrell, and Rohrbach}]{park2018multimodal}
Dong~Huk Park, Lisa~Anne Hendricks, Zeynep Akata, Anna Rohrbach, Bernt Schiele, Trevor Darrell, and Marcus Rohrbach. 2018.
\newblock Multimodal explanations: Justifying decisions and pointing to the evidence.
\newblock In \emph{Proceedings of the IEEE conference on computer vision and pattern recognition}, pages 8779--8788.

\bibitem[{Pruthi et~al.(2022)Pruthi, Bansal, Dhingra, Soares, Collins, Lipton, Neubig, and Cohen}]{pruthi2022evaluating}
Danish Pruthi, Rachit Bansal, Bhuwan Dhingra, Livio~Baldini Soares, Michael Collins, Zachary~C Lipton, Graham Neubig, and William~W Cohen. 2022.
\newblock Evaluating explanations: How much do explanations from the teacher aid students?
\newblock \emph{Transactions of the Association for Computational Linguistics}, 10:359--375.

\bibitem[{Rafailov et~al.(2024)Rafailov, Sharma, Mitchell, Manning, Ermon, and Finn}]{rafailov2024direct}
Rafael Rafailov, Archit Sharma, Eric Mitchell, Christopher~D Manning, Stefano Ermon, and Chelsea Finn. 2024.
\newblock Direct preference optimization: Your language model is secretly a reward model.
\newblock \emph{Advances in Neural Information Processing Systems}, 36.

\bibitem[{Rajani et~al.(2019)Rajani, McCann, Xiong, and Socher}]{rajani-etal-2019-explain}
Nazneen~Fatema Rajani, Bryan McCann, Caiming Xiong, and Richard Socher. 2019.
\newblock \href {https://doi.org/10.18653/v1/P19-1487} {Explain yourself! leveraging language models for commonsense reasoning}.
\newblock In \emph{Proceedings of the 57th Annual Meeting of the Association for Computational Linguistics}, pages 4932--4942, Florence, Italy. Association for Computational Linguistics.

\bibitem[{Ramnath et~al.(2024)Ramnath, Joshi, Hallinan, Lu, Li, Chan, Hessel, Choi, and Ren}]{ramnath2024tailoring}
Sahana Ramnath, Brihi Joshi, Skyler Hallinan, Ximing Lu, Liunian~Harold Li, Aaron Chan, Jack Hessel, Yejin Choi, and Xiang Ren. 2024.
\newblock \href {https://openreview.net/forum?id=t8eO0CiZJV} {Tailoring self-rationalizers with multi-reward distillation}.
\newblock In \emph{The Twelfth International Conference on Learning Representations}.

\bibitem[{Ribeiro et~al.(2016)Ribeiro, Singh, and Guestrin}]{ribeiro2016should}
Marco~Tulio Ribeiro, Sameer Singh, and Carlos Guestrin. 2016.
\newblock "{Why should I trust you?}" explaining the predictions of any classifier.
\newblock In \emph{Proceedings of the 22nd ACM SIGKDD international conference on knowledge discovery and data mining}, pages 1135--1144.

\bibitem[{Samek et~al.(2016)Samek, Binder, Montavon, Lapuschkin, and M{\"u}ller}]{samek2016evaluating}
Wojciech Samek, Alexander Binder, Gr{\'e}goire Montavon, Sebastian Lapuschkin, and Klaus-Robert M{\"u}ller. 2016.
\newblock Evaluating the visualization of what a deep neural network has learned.
\newblock \emph{IEEE transactions on neural networks and learning systems}, 28(11):2660--2673.

\bibitem[{Selvaraju et~al.(2017)Selvaraju, Cogswell, Das, Vedantam, Parikh, and Batra}]{selvaraju2017grad}
Ramprasaath~R Selvaraju, Michael Cogswell, Abhishek Das, Ramakrishna Vedantam, Devi Parikh, and Dhruv Batra. 2017.
\newblock Grad-cam: Visual explanations from deep networks via gradient-based localization.
\newblock In \emph{Proceedings of the IEEE international conference on computer vision}, pages 618--626.

\bibitem[{Touvron et~al.(2023)Touvron, Martin, Stone, Albert, Almahairi, Babaei, Bashlykov, Batra, Bhargava, Bhosale et~al.}]{touvron2023llama}
Hugo Touvron, Louis Martin, Kevin Stone, Peter Albert, Amjad Almahairi, Yasmine Babaei, Nikolay Bashlykov, Soumya Batra, Prajjwal Bhargava, Shruti Bhosale, et~al. 2023.
\newblock Llama 2: Open foundation and fine-tuned chat models.
\newblock \emph{arXiv preprint arXiv:2307.09288}.

\bibitem[{Turpin et~al.(2024)Turpin, Michael, Perez, and Bowman}]{turpin2024language}
Miles Turpin, Julian Michael, Ethan Perez, and Samuel Bowman. 2024.
\newblock Language models don't always say what they think: Unfaithful explanations in chain-of-thought prompting.
\newblock \emph{Advances in Neural Information Processing Systems}, 36.

\bibitem[{Wallace et~al.(2018)Wallace, Feng, and Boyd-Graber}]{Wallace2018InterpretingNN}
Eric Wallace, Shi Feng, and Jordan~L. Boyd-Graber. 2018.
\newblock \href {https://api.semanticscholar.org/CorpusID:52182354} {Interpreting neural networks with nearest neighbors}.
\newblock In \emph{BlackboxNLP@EMNLP}.

\bibitem[{Wang et~al.(2022)Wang, Wei, Schuurmans, Le, Chi, Narang, Chowdhery, and Zhou}]{wang2022self}
Xuezhi Wang, Jason Wei, Dale Schuurmans, Quoc Le, Ed~Chi, Sharan Narang, Aakanksha Chowdhery, and Denny Zhou. 2022.
\newblock Self-consistency improves chain of thought reasoning in language models.
\newblock \emph{arXiv preprint arXiv:2203.11171}.

\bibitem[{Wei et~al.(2022)Wei, Wang, Schuurmans, Bosma, Xia, Chi, Le, Zhou et~al.}]{wei2022chain}
Jason Wei, Xuezhi Wang, Dale Schuurmans, Maarten Bosma, Fei Xia, Ed~Chi, Quoc~V Le, Denny Zhou, et~al. 2022.
\newblock Chain-of-thought prompting elicits reasoning in large language models.
\newblock \emph{Advances in neural information processing systems}, 35:24824--24837.

\bibitem[{Wiegreffe et~al.(2022)Wiegreffe, Hessel, Swayamdipta, Riedl, and Choi}]{wiegreffe-etal-2022-reframing}
Sarah Wiegreffe, Jack Hessel, Swabha Swayamdipta, Mark Riedl, and Yejin Choi. 2022.
\newblock \href {https://doi.org/10.18653/v1/2022.naacl-main.47} {Reframing human-{AI} collaboration for generating free-text explanations}.
\newblock In \emph{Proceedings of the 2022 Conference of the North American Chapter of the Association for Computational Linguistics: Human Language Technologies}, pages 632--658, Seattle, United States. Association for Computational Linguistics.

\bibitem[{Woodward and Ross(2003)}]{woodward2003scientific}
JF~Woodward and L~Ross. 2003.
\newblock Scientific explanation: Stanford encyclopedia of philosophy.
\newblock \emph{Summer 2021 Edition}.

\bibitem[{Wu and Mooney(2019)}]{wu-mooney-2019-faithful}
Jialin Wu and Raymond Mooney. 2019.
\newblock \href {https://doi.org/10.18653/v1/W19-4812} {Faithful multimodal explanation for visual question answering}.
\newblock In \emph{Proceedings of the 2019 ACL Workshop BlackboxNLP: Analyzing and Interpreting Neural Networks for NLP}, pages 103--112, Florence, Italy. Association for Computational Linguistics.

\bibitem[{Ye and Durrett(2022)}]{ye2022unreliability}
Xi~Ye and Greg Durrett. 2022.
\newblock The unreliability of explanations in few-shot prompting for textual reasoning.
\newblock \emph{Advances in neural information processing systems}, 35:30378--30392.

\bibitem[{Young et~al.(2024)Young, Chen, Li, Huang, Zhang, Zhang, Wang, Li, Zhu, Chen et~al.}]{young2024yi}
Alex Young, Bei Chen, Chao Li, Chengen Huang, Ge~Zhang, Guanwei Zhang, Guoyin Wang, Heng Li, Jiangcheng Zhu, Jianqun Chen, et~al. 2024.
\newblock Yi: Open foundation models by 01. ai.
\newblock \emph{arXiv preprint arXiv:2403.04652}.

\bibitem[{Zhou et~al.(2022{\natexlab{a}})Zhou, Sch{\"a}rli, Hou, Wei, Scales, Wang, Schuurmans, Cui, Bousquet, Le et~al.}]{zhou2022least}
Denny Zhou, Nathanael Sch{\"a}rli, Le~Hou, Jason Wei, Nathan Scales, Xuezhi Wang, Dale Schuurmans, Claire Cui, Olivier Bousquet, Quoc Le, et~al. 2022{\natexlab{a}}.
\newblock Least-to-most prompting enables complex reasoning in large language models.
\newblock \emph{arXiv preprint arXiv:2205.10625}.

\bibitem[{Zhou et~al.(2023)Zhou, Zhang, and Tan}]{zhou-etal-2023-flame}
Yangqiaoyu Zhou, Yiming Zhang, and Chenhao Tan. 2023.
\newblock \href {https://doi.org/10.18653/v1/2023.acl-long.372} {{FL}am{E}: Few-shot learning from natural language explanations}.
\newblock In \emph{Proceedings of the 61st Annual Meeting of the Association for Computational Linguistics (Volume 1: Long Papers)}, pages 6743--6763, Toronto, Canada. Association for Computational Linguistics.

\bibitem[{Zhou et~al.(2022{\natexlab{b}})Zhou, Booth, Ribeiro, and Shah}]{zhou2022feature}
Yilun Zhou, Serena Booth, Marco~Tulio Ribeiro, and Julie Shah. 2022{\natexlab{b}}.
\newblock Do feature attribution methods correctly attribute features?
\newblock In \emph{Proceedings of the AAAI Conference on Artificial Intelligence}, volume~36, pages 9623--9633.

\end{thebibliography}

\clearpage

\appendix

\section{Appendices}
\subsection{Explanation Prompting}
\label{app:explanation_prompting}

We use the following GPT-4 generated explanation as a one-shot prompt to guide models in explaining the TripAdvisor hotel review dataset (see \autoref{sec:problem}):

\noindent\fbox{%
    \parbox{\columnwidth}{%
    \small
\textbf{Review:} \textpr{We stayed at Hotel Monaco and upgraded to a suite. I was disappointed with the room layout, and shabby furniture. I bumped into the coffee table and the top almost crashed to the ground. The make up mirror in the bathroom didnt work, the TV reception was very poor and overall the room just wasnt worth the \$450 we spent there. Next time, we will go back to Trump Tower across the street, which is far superior to Hotel Monaco.}

\textbf{Explanation for truthful prediction:} \textpr{[reason1] Suggests an alternative hotel: The reviewer suggests an alternative hotel (Trump Tower) that they believe is superior, which is a practical and often truthful element in a hotel review. [reason2] Mentions specific details: The reviewer mentions specific details about the room, such as the room layout, shabby furniture.}

\textbf{Review:} \textpr{My stay at the Hotel Monaco recently was fantastic. We got a very reasonable price for our stay (especially since we were a AAA member), and there was plenty of perks to utilize; in-room spa treatments and an evening wine wine hour to name a few. Another thing that the kids really enjoyed was that, upon request, the staff brought us a fish in its own fishbowl for our room. How cool is that! The staff were all very helpful and friendly and our stay was very enjoyable. I would definitely recommend the Hotel Monaco to anyone in the downtown Chicago area.}

\textbf{Explanation for deceptive prediction:} \textpr{[reason1] Generic and overly positive tone: The review maintains a very positive tone without any specific details or occurrences that would substantiate such enthusiasm. The language used is quite general, lacking specific details or instances that typically accompany genuine experiences. [reason2] Uncommon or unrealistic details: The mention of receiving a fish in a fishbowl upon request is an uncommon and specific detail. While some hotels do offer unique services, the presentation of this detail, without further elaboration or context, seems exaggerated and implausible, casting doubt on the review's authenticity.}
    }%
} \newline

To guide models in generating explanations for the Amazon product review dataset (see \autoref{sec:problem}), we use the following GPT-4 generated explanation as a one-shot example:

\noindent\fbox{%
    \parbox{0.96\columnwidth}{%
    \small
\textbf{Review:} \textpr{I only wanted to use this to charge two phones from one adapter.  This item at first seemed to be of decent quality.  However, after 3 months of occasional use, the casing of one of the male plugs just fell apart in my hand as I was plugging it in. I did not apply undue force when it happened. After closer inspection, I could see how weak the plastic was -- tiny, brittle plastic clips hold the case together. I might have been able to glue or tape it back together, but given that it charges \$500 phones, not worth the risk of a short.  There is a diode and and two wires soldered inside the housing that broke that could easilly short circuit.  I do not recommend this product if you value the items you are charging.}

\textbf{Explanation for truthful prediction:} \textpr{[reason1] Firsthand experience: The reviewer shares a personal account of using the product over three months, offering a credible, real-world perspective rather than speculation. [reason2] Safety concerns explained logically: The reviewer points out that the internal wiring and diode were exposed after the casing broke, increasing the risk of a short circuit.}

\textbf{Review:} \textpr{I was looking to downsize from an otterbox defender (whose holster's clip failed after one year) and this fits the bill nicely. Unlike the defender, the outside of the case is hard so it slides around (and in and out of your pocket) easily. You also don't have to open and close littler rubber doors to gain access to the headphone jack and charging/syncing port. The only hassle is that the rubber lining around the headphone jack wouldn't allow some of my older cheap cords or my good sony MDR 7506 headphones to plug in. Trimming the rubber solved the cheap cord problems but I had to replace the cord on the sonys with a hacked media bridge cord with a stepped plug on it.}

\textbf{Explanation for deceptive prediction:} \textpr{[reason1] Minimizes major flaws: The review glosses over a serious issue — the case blocks headphone and charging ports — by treating it as a minor fix. [reason2] Shifts blame: Suggests the problem is due to 'older cheap cords' rather than a design flaw, misleading readers.}
    }%
}

\subsection{Sensitivity to Prompt Format}
\label{app:prompt_format}

To assess whether the PEX consistency score is sensitive to different prompt formats, we use 3 different prompt formats to compute the PEX consistency score described in \autoref{eq:adjusted_woe}, where $\explanation$ is a free-text explanation to be measured:

\noindent V1: $\prompt'_1 (\question, \explanation)$ = \textpr{Is this review truthful or deceptive? Review: \{review\}. Answer: $\explanation$. Therefore, this review is}

\noindent V2: $\prompt'_2 (\question, \explanation)$ = \textpr{Please classify the following review as either truthful or deceptive: \{review\}. Answer: $\explanation$. Hence, this review is}

\noindent V3: $\prompt'_3 (\question, \explanation)$ = \textpr{Evaluate the authenticity of this review: \{review\}. Question: Is the review truthful or deceptive? Analysis:  $\explanation$. Answer: This review is}

We then compute Kendall rank correlations between PEX consistency scores obtained from each pair of prompt versions. The results are shown in \autoref{tab:sensitivity}, indicating PEX scores are relatively robust to variations in prompt format.

\subsection{Supervised Fine-tuning and Student Model Training}
\label{app:sft}

For supervised fine-tuning (SFT), we finetune Llama-2-13b-chat \cite{touvron2023llama}, Mistral-7B-Instruct-v0.3 \cite{mistral7b-instruct-v0.3} and Yi-1.5-9B-chat \cite{young2024yi} models to improve their prediction accuracy on two opinion spam classification datasets (\autoref{sec:consistency}).

\paragraph{Hyperparameters and computation.}

For Llama-2, Mistral and Yi-1.5 models, we finetune our models for $20$ epochs, using a learning rate of $2e^{-4}$, batch size of $2$, and AdamW optimizer. We use a LoRA adapter~\cite{hu2021lora} of rank $64$ and alpha $16$. We finetune the models using $1$ RTXA6000 GPU, for $\sim 24$ hours.
We train the student models for $100$ epochs using $10$ training examples and $50$ epochs using $20$ training examples, due to a small number of training samples. Training each model takes 1-2 hours.

\paragraph{Tools.}
We implement our models with Pytorch 2.0.1, Huggingface Transformers 4.31.0, scikit-learn 1.2.2 and SciPy 1.15.1.

\paragraph{Datasets.}
We two opinion spam detection datasets: (i) TripAdvisor hotel review dataset \cite{ott2013negative} and 
(ii) Amazon product review dataset \cite{hussain2020spam}, under their Creative Commons Attribution-NonCommercial-ShareAlike 3.0 Unported License.

\begin{table*}[!htbp]
\centering
\small
\renewcommand{\arraystretch}{1.2} 
\setlength{\tabcolsep}{5pt} 
\begin{tabular}{@{}lccccccccc@{}}
\toprule
\textbf{Model} & \multicolumn{3}{c}{\textbf{Mistral}} & \multicolumn{3}{c}{\textbf{Llama-2}} & \multicolumn{3}{c}{\textbf{Yi-1.5}} \\
\cmidrule(lr){2-4} \cmidrule(lr){5-7} \cmidrule(lr){8-10}
& \textbf{$k$=10} & \textbf{$k$=20} & \textbf{Avg} & \textbf{$k$=10} & \textbf{$k$=20} & \textbf{Avg} & \textbf{$k$=10} & \textbf{$k$=20} & \textbf{Avg} \\
\midrule
\textbf{Pred Only}  & 44.0±7.8 & 66.9±1.5 & 55.5 & 54.7±0.1 & 64.2±1.4 & 59.6 & 62.4±0.0 & 63.6±0.0 & 63.0 \\ \midrule
\textbf{+ SFT}  & 64.2±1.8 & 68.1±0.2 & 66.2 & 56.0±4.0 & 60.6±2.2 & 58.3 & 66.5±0.0 & 65.4±0.7 & 66.0 \\
\textbf{+ DPO}  & \textbf{69.1±1.2} & \textbf{70.4±1.0} & \textbf{69.8} & \textbf{61.4±0.9} & \textbf{65.3±2.9} & \textbf{63.4} & \textbf{69.3±0.1} & \textbf{72.5±0.1} & \textbf{70.9} \\
\bottomrule
\end{tabular}
\caption{Simulation performance (F1) of student models on the TripAdvisor test set, evaluating how well they approximate the teacher model’s predictions on unseen examples. The student models are trained using explanations from different teacher models: supervised fine-tuning (SFT) or direct preference optimization (DPO) with \ourname consistency measure. The variable $k$ represents the size of the teacher model's training samples. The error bars after ± represent 95\% confidence intervals across 5 training passes.}
\label{tab:tripadvisor_eval}
\end{table*}

\begin{table*}[!htbp]
\centering
\small
\renewcommand{\arraystretch}{1.2} 
\setlength{\tabcolsep}{5pt} 
\begin{tabular}{@{}lccccccccc@{}}
\toprule
\textbf{Model} & \multicolumn{3}{c}{\textbf{Mistral}} & \multicolumn{3}{c}{\textbf{Llama-2}} & \multicolumn{3}{c}{\textbf{Yi-1.5}} \\
\cmidrule(lr){2-4} \cmidrule(lr){5-7} \cmidrule(lr){8-10}
& \textbf{$k$=10} & \textbf{$k$=20} & \textbf{Avg} & \textbf{$k$=10} & \textbf{$k$=20} & \textbf{Avg} & \textbf{$k$=10} & \textbf{$k$=20} & \textbf{Avg} \\
\midrule
\textbf{Pred Only}  & 78.9±0.8 & 87.4±0.4 & 83.2 & 85.0±1.5 & 92.2±0.9 & 88.6 & 53.8±11.7 & 81.8±1.8 & 67.8 \\ \midrule
\textbf{+ SFT}  & 83.6±0.4 & 86.8±0.3 & 85.2 & 85.6±2.8 & 92.8±1.8 & 89.2 & 68.0±11.8 & 82.9±2.7 & 75.5 \\
\textbf{+ DPO}  & \textbf{84.0±0.5} & \textbf{89.3±0.7} & \textbf{86.7} & \textbf{89.9±1.6} & \textbf{93.2±1.0} & \textbf{91.6} & \textbf{81.9±8.1} & \textbf{88.4±3.1} & \textbf{85.2} \\
\bottomrule
\end{tabular}
\caption{Simulation performance (F1) of student models on the Amazon test set, evaluating how well they approximate the teacher model’s predictions on unseen examples. The student models are trained using explanations from different teacher models: supervised fine-tuning (SFT) or direct preference optimization (DPO) with \ourname consistency measure. The variable $k$ represents the size of the teacher model's training samples. The error bars after ± represent 95\% confidence intervals across 5 training passes.}
\label{tab:amazon_eval}
\end{table*}

\subsection{Direct Preference Optimization}
\label{app:dpo}

\paragraph{Hyperparameters and computation.}

For Llama-2, we finetune our models using a learning rate of $2e^{-4}$. For Mistral and Yi-1.5, we finetune our models using a learning rate of $2e^{-5}$.
We finetune all the models for $5$ epochs on the TripAdvisor dataset, and $1$ epoch on the Amazon dataset. We use batch size of $2$, AdamW optimizer, a LoRA adapter~\cite{hu2021lora} of rank $32$ and alpha $16$. We finetune the models using $1$ RTXA6000 GPU, for $\sim 24$ hours.

\paragraph{Tools.} We implement DPO training using TRL (Transformer Reinforcement Learning) 0.14.0.

\subsection{Explanation Faithfulness Evaluation}
\label{app:eval}

\autoref{tab:tripadvisor_eval} and \autoref{tab:amazon_eval} show the simulation performance for explanation faithfulness evaluation,  broken down by different values of $k$, which represents the number of training samples being used to train the student model.

\begin{figure*}[h]
    \centering
    \includegraphics[width=\textwidth]{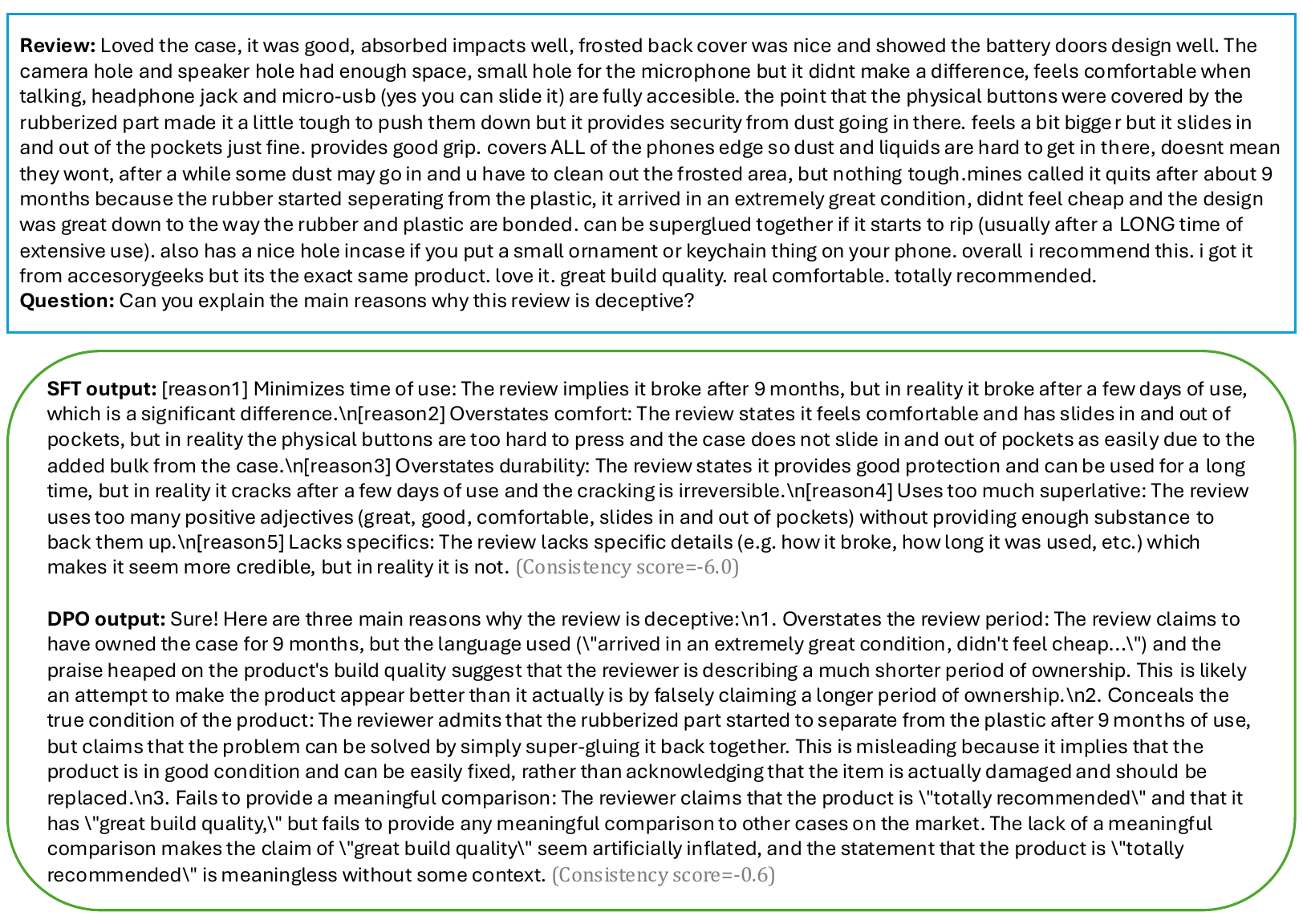} 
    \caption{
     DPO (Direct Preference Optimization) improves consistency score from the Llama-2 SFT (Supervised Fine-Tuning) model, but the generated explanation is still not consistent enough for its deceptive prediction on an Amazon product review.}
    \label{fig:qual_example2}
\end{figure*}

\subsection{Additional Qualitative Examples}
\label{app:qualitative}

\autoref{fig:qual_example2} shows additional qualitative examples for the explanations generated by DPO and SFT.

\end{document}